\documentclass[runningheads]{llncs}

\usepackage[noend]{algpseudocode} 
\usepackage{algorithm}
\usepackage{amssymb}
\usepackage[inline,shortlabels]{enumitem}
\usepackage{graphicx}
\usepackage{url}  
\usepackage{todonotes}
\usepackage{listings}
\usepackage[T1]{fontenc}
\usepackage[utf8]{inputenc}
\usepackage{tikz}
\usetikzlibrary{arrows,fit, calc, positioning, automata, shapes.geometric}
\usepackage{hyperref}
\usepackage{verbatim}
\usepackage{subfig}
\usepackage[gen]{eurosym}
\usepackage{enumitem}

\colorlet{punct}{red!60!black}
\definecolor{background}{HTML}{EEEEEE}
\definecolor{delim}{RGB}{20,105,176}
\colorlet{numb}{magenta!60!black}

\lstdefinelanguage{json}{
    basicstyle=\normalfont\ttfamily,
    numbers=left,
    numberstyle=\scriptsize,
    stepnumber=1,
    numbersep=8pt,
    showstringspaces=false,
    breaklines=true,
    frame=lines,
    backgroundcolor=\color{background},
    literate=
     *{0}{{{\color{numb}0}}}{1}
      {1}{{{\color{numb}1}}}{1}
      {2}{{{\color{numb}2}}}{1}
      {3}{{{\color{numb}3}}}{1}
      {4}{{{\color{numb}4}}}{1}
      {5}{{{\color{numb}5}}}{1}
      {6}{{{\color{numb}6}}}{1}
      {7}{{{\color{numb}7}}}{1}
      {8}{{{\color{numb}8}}}{1}
      {9}{{{\color{numb}9}}}{1}
      {:}{{{\color{punct}{:}}}}{1}
      {,}{{{\color{punct}{,}}}}{1}
      {\{}{{{\color{delim}{\{}}}}{1}
      {\}}{{{\color{delim}{\}}}}}{1}
      {[}{{{\color{delim}{[}}}}{1}
      {\$}{{{\color{numb}{\$}}}}{1}
      {]}{{{\color{delim}{]}}}}{1},
}
\definecolor{dkgreen}{rgb}{0,0.6,0}

\lstdefinelanguage{sparql}{
    basicstyle=\normalfont\ttfamily,
    numbers=left,
    numberstyle=\scriptsize,
    stepnumber=1,
    numbersep=8pt,
    showstringspaces=false,
    breaklines=true,
    frame=lines,
    comment=[l]{\#},
    commentstyle=\color{purple}\ttfamily,
    stringstyle=\color{red}\ttfamily,
    morestring=[b]",
    keywordstyle=\color{dkgreen},
    morekeywords={id,v1,v2,v3r,v31},
    backgroundcolor=\color{background},
    literate=
      {\?}{{{\color{dkgreen}{?}}}}{1}
}

\begin{document}
%
\title{CLARIAH: Enabling Interoperability Between Humanities Disciplines with Ontologies}
%
%
\author{Albert Mero\~{n}o-Pe\~{n}uela\inst{1,3}
\and Victor de Boer\inst{1}
\and Marieke van Erp\inst{2}
\and Richard Zijdeman\inst{2,3,4}
\and Rick Mourits\inst{5}
\and Willem Melder\inst{6}
\and Auke Rijpma\inst{5}
\and Ruben Schalk\inst{5}
}

%
\authorrunning{A. Meroño-Peñuela \textit{et al.}}
\titlerunning{CLARIAH: Humanities Interoperability}

\institute{Vrije Universiteit, Amsterdam, The Netherlands 
\and
KNAW Humanities Cluster, DHLab, Amsterdam, The Netherlands
\and 
International Institute of Social History, Amsterdam, The Netherlands
\and
University of Stirling, Stirling, Scotland
\and
Utrecht University, Utrecht, The Netherlands
\and
Netherlands Institute for Sound and Vision, Hilversum, The Netherlands
}

\maketitle 

\setcounter{footnote}{0}

\begin{abstract}
One of the most important goals of digital humanities is to provide researchers with data and tools for new research questions, either by increasing the scale of scholarly studies, linking existing databases, or improving the accessibility of data. 
Here, the FAIR principles provide a useful framework.
Integrating data from diverse humanities domains is not trivial, research questions such as “was economic wealth equally distributed in the 18th century?”, or “what are narratives constructed around disruptive media events?”) and preparation phases (e.g. data collection, knowledge organisation, cleaning) of scholars need to be taken into account. In this chapter, we describe the ontologies and tools developed and integrated in the Dutch national project CLARIAH to address these issues across datasets from three fundamental domains or “pillars” of the humanities (linguistics, social and economic history, and media studies) that have paradigmatic data representations (textual corpora, structured data, and multimedia). We summarise the lessons learnt from using such ontologies and tools in these domains from a generalisation and reusability perspective.

\end{abstract}

\section{Introduction}
\label{sec:introduction}

The digital humanities (DH) are a “movement and a push to apply the tools and methods of computing to the subject matter to the humanities” \cite{haigh2014we}. By increasing the availability of digital data and compute power, the DH, ultimately, strives to attain broader insights in settled debates and allow for the study of new questions \cite{renckens2016digital}. This mimics the approach of “big data” that has been successful in other scientific fields over the past decades \cite{collins2003human}.

However, DH research is currently very challenging, as the reusability of humanities datasets is limited \cite{ashkpour2019theory,hoekstra2018datalegend}, due to their low fulfilment of the FAIR data principles for scientific data management \cite{wilkinson2016fair}.
These state that data needs to be FAIR \cite{wilkinson2016fair}.
Findability is problematic because humanities datasets are typically scattered, small, and represented in non-machine readable ways (for example, scanned images), lack globally unique and persistent identifiers, rich metadata, and registration in search indexes. Accessibility is limited as data are often in possession of a handful of individual researchers; these datasets are, therefore, hard to access automatically through open protocols. Reusability, greatly empowered by releasing data under open access licenses (e.g. creative commons), is hampered by a research culture that does not totally embrace open licensing yet. But, most importantly, interoperability among these datasets is severely compromised due to the scarce use of formal knowledge representation languages, shared vocabularies, and ontologies. Linked Data and Semantic Web technologies have proven to be effective at addressing these issues in various domains. For example, Bioportal \cite{noy2009bioportal} is a comprehensive repository of ontologies to address interoperability in biomedicine. Similar successes have been observed in cross-domain, geographic, government, media and library datasets \cite{heath2011linked}.

In the Netherlands, we are taking up this interoperability mission in CLARIAH, by developing a nation-wide, common data space for the Arts and the Humanities. The main objectives of CLARIAH are: (a) facilitating the publication and reuse of humanities data following the FAIR principles and Linked Data standards
(we currently serve DH researchers with 114 datasets containing 865 billion RDF triples in 1,500 Knowledge Graphs); and (b) to do so by fostering the collaboration between different research disciplines (historians, social scientists, linguists, media studies, and computer scientists) that had rarely collaborated before at this scale. 
CLARIAH consists of three pillars, each combining a domain with a technical challenge. The \textbf{economic and social history pillar} deals with structured data. The field has a long tradition of creating tabular datasets and the challenge is to gather and integrate these datasets, both within and outside of the domain. The \textbf{linguistic pillar} aims to provide research infrastructure facilities for carrying out linguistic research. This includes tools, and resources to support the study of language as well as to support automatic text analysis tools for other domains.  The \textbf{media studies pillar} deals with audiovisual data, and aims to provide infrastructure for researchers that study and annotate mass media, newspapers, film, radio, television, and their contexts, and the central role these have played in the emergence of modern societies.

This publication of reusable humanities datasets through interdisciplinary collaboration poses, however, a more important challenge: to find adequate ontology engineering practices and social processes to reach agreement in what and how ontologies and vocabularies can be used to enable quantitative and comparative analyses in the humanities. 
In this chapter we address the questions: how can ontologies and Semantic Web technologies enable 
data interoperability
in the humanities?
What ontology engineering practices are effective in enabling cooperation in research communities that have, traditionally, been distant? We do this through the lens of (a) Semantic Web ontologies that we have engineered in order to model key domains in the humanities; and (b) a set of tools that we have developed to leverage these ontologies as data interoperability enablers. Combined, these two insights help to streamline the  fundamental phases of humanities data preparation (e.g. data collection, knowledge organisation, cleaning), which are typically the most time-consuming tasks in research projects \cite{garijo2014common}. Specifically, our contributions are:

\begin{itemize}[topsep=0pt]
\item A set of ontologies and tools enabling data interoperability in social and economic history, language, and media studies (Sections \ref{sec:ontologies-wp4}, \ref{sec:ontologies-wp3}, \ref{sec:ontologies-wp5})
\item A description of our efforts to use these ontologies in a fragmented landscape, increasing their coupling and reusability
(Section \ref{sec:ontologies-inter})
\item Guidelines and lessons learned on large institutional ontology engineering efforts while developing all these (Section \ref{sec:conclusion})
\end{itemize}

\section{Related work}
\label{sec:related-work}

We survey related work in ontologies and tools for the different CLARIAH pillars.

In the area of economic and social history and structured data,
historical ontologies \cite{merono2015semantic} are relevant, but their focus historical narratives and events rather than registry-based datasets, typical of the domain, make them generally unfit.
The W3C standards CSV on the Web \cite{world2016csv} and 
RDF Data Cube \cite{world2014rdf} are better suited for historical tabular and multidimensional data; albeit agnostic with respect to specific domain concepts. These are typically used to describe ordinary people, such as their occupations (e.g. ISCO-88 \cite{elias1997occupational}), life events (e.g. Bio vocabulary \cite{davis2004bio}) and general metadata (e.g. Schema.org \cite{guha2016schema}). However, none of these standards specifically cover the historical nuances. Previous work addressing this either focuses on record linkage \cite{mandemakers2017links} or technical ecosystems \cite{hoekstra2018datalegend}. Here, we focus on the  shared ontological models that these methods need to interoperate.

Relevant to the linguistics pillar are the various standards to express text enrichments, such as multiple tagsets for e.g. part-of-speech tagging \cite{petrov2011universal} and lexicons \cite{maks2016integrating}. Many well-established standards for annotations (e.g. W3C  Web Annotation Model \cite{sanderson2013open}; the NAF\footnote{\url{http://wordpress.let.vupr.nl/naf/}} and FoLiA formats \cite{gompel2017folia}) and lexicon models for ontologies (e.g. Lemon \cite{mccrae2017ontolex}) exist. Despite the extensibility of these standards, they usually lack concepts necessary to describe the temporal scope of a word sense, or simply do not interoperate.
In our work, we focus more on providing shared means to work across these, and extending them where possible towards the diachronic aspect, rather than reconciling them.

Various existing ontologies are relevant for audiovisual archiving and media studies. For example, the Simple Event Model (SEM) \cite{van2011design} captures the time and place aspects of a domain, and models events as complex relations between people, places, actions and objects. These are typical properties in the metadata of audiovisual material, many times described with CIDOC \cite{doerr2003cidoc} (for museum assets) and FRBR \cite{tillett2005frbr} (for library assets such as hierarchical works, manifestations and realisations). Other general models for metadata, such as SKOS \cite{miles2005skos}, Dublin Core \cite{weibel1998dublin} exist, but even in concrete models for cultural heritage (e.g. Europeana Data Model \cite{doerr2010europeana}) none specifically targets Dutch audiovisual archives.




More recently, the workshops series on Humanities in the Semantic Web (WHiSe) have been organised as satellite events in Semantic Web conferences \cite{adamou2016whise,adamou2017whise}; 
and some methodologies (e.g. Ontology Design Patterns \cite{gangemi2009ontology}; FAIR \cite{wilkinson2016fair}) directly address reusability of models for interoperability.
Overall, no existing approach gathers all ontologies specifically developed for humanities data. Here, we describe how in CLARIAH we contribute to this growing body of ontologies for the humanities.

\section{Ontologies in CLARIAH}
\label{sec:ontologies}

In this section, we describe the ontologies developed in CLARIAH for the purpose of fulfilling the project goals regarding data interoperability. We focus on ontologies and tools for each of the ``pillars'' (Sections \ref{sec:ontologies-wp4}, \ref{sec:ontologies-wp3}, \ref{sec:ontologies-wp5}) as well as their interactions and connections (Section \ref{sec:ontologies-inter}).

\subsection{Social Economic History \& Structured Data}
\label{sec:ontologies-wp4}

Within 
social and economic history, quantitative analysis of registers and ledgers - among others - are extensively used to study the history of ordinary people \cite{ruggles2019history}. 
To integrate and combine these datasets, we describe here the ontologies and tools 
we assembled. 

\subsubsection{Ontologies}
\label{sec:ontologies-wp4-onto}

\begin{figure}[ht!]
    \centering
    \includegraphics[width=\textwidth]{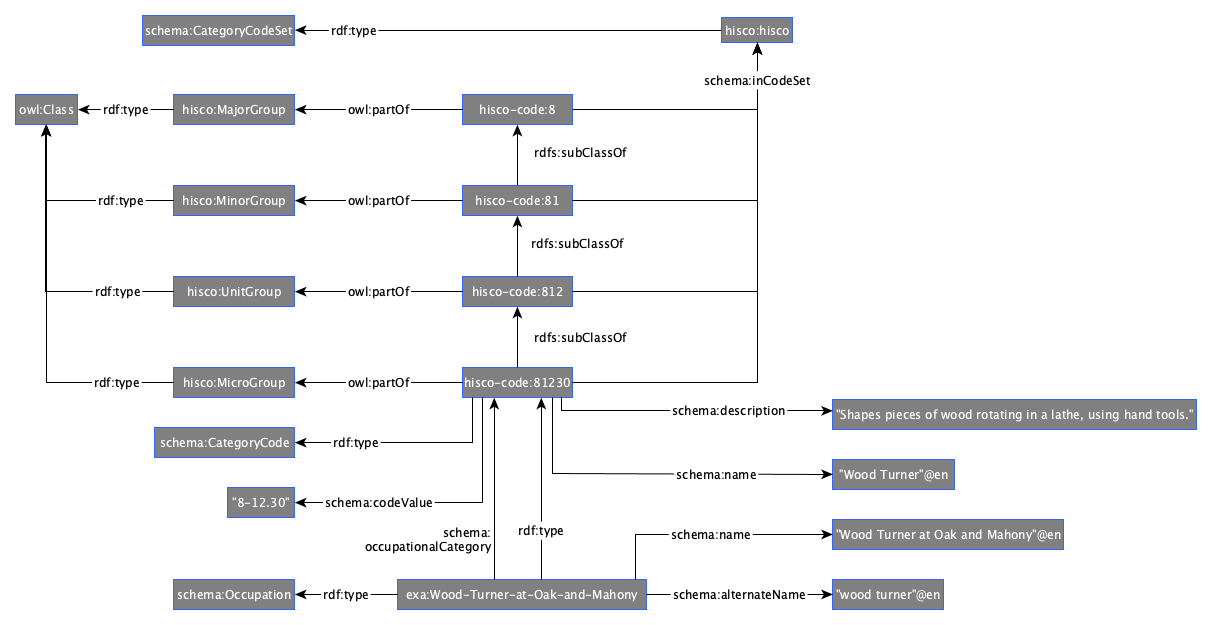}
    \caption{HISCO schema for an example occupation title, using the \texttt{hisco}, \texttt{schema}, \texttt{rdfs}, \texttt{owl} namespaces.}
    \label{fig:hisco}
\end{figure}

Most historical quantitative datasets are indexes of official, institutional records of the past, such as national censuses, statistical national offices, and tax registers. 
For example, when studying the \textbf{history of work} occupations are one of the key indicators in historical inequality research as it is commonly available for multiple centuries and in sources across the globe. To appreciate differences within occupations over time and across the globe international standards have been devised. The Historical International Standard Classification of Occupations (HISCO) \cite{leeuwen2002hisco} categorises occupations by the activities of their incumbents, while the HIStorical CAMbridge Scale (HISCAM) \cite{lambert2013construction} orders those categories based on the incumbents social interactions. 
We created multiple versions of a vocabulary to use HISCO and HISCAM as Linked Data. 
The first version was very much oriented towards SDMX Occupation ISCO-88 \cite{elias1997occupational} value list, but this work appears to have been abandoned. 
This SKOS oriented version allows for easy referencing towards multiple layers in HISCO and HISCAM using the \texttt{skos:broader} and \texttt{skos:broaderTransitive} relations and was attractive, because ISCO-88 could be seen as the contemporary equivalent of HISCO and thus modeling them in the same way made sense. Meanwhile there were discussions with the Schema.org community \cite{guha2016schema} on how to model occupational schema generically, while allowing for occupational schema specificity. This has led to the latest version of the HISCO and HISCAM schema (see Figure \ref{fig:hisco}), which are fully represented via Schema.org. To describe the hierarchical structure of HISCO, we additionally borrowed from RDFS and OWL.

Similarly, many historical records and registers, such as censuses, describe the \textbf{historical religious affiliations} of individuals and groups.
Throughout the centuries, many different religious denominations have appeared. However, these denominations are represented in the original datasets in different ways and levels of detail, e.g. Confucianism and Ruism, or the various subcategories of Islam (Sunni, Shia, etc.) or Christianity (Catholic, Protestant, Orthodox, etc.). These changing denominations pose a challenge to researchers in the field, who need unambiguous ways of defining religious denominations for comparative analyses.
To address this, we developed the Linked International Classification for Religions\footnote{See \url{https://datasets.iisg.amsterdam/dataverse/LICR}} (LICR), a SKOS \cite{miles2005skos} taxonomy that organises historical religious denominations into 131 categories and subcategories; and provides mappings to various other well known religious classification systems, such as IPUMS, NAPP, and HL7. These mappings are expert-made.
In addition, LICR is enriched with links to equivalent DBpedia resources
LICR is built on top of (1) the unique religion identifiers of these systems, adding unique identifiers of its own; and (2) the Linked Data design principles \cite{heath2011linked}, therefore producing a unique dereferenceable URI for each of these identifiers. 

\begin{figure}[ht]
    \centering
    \includegraphics[width=\textwidth]{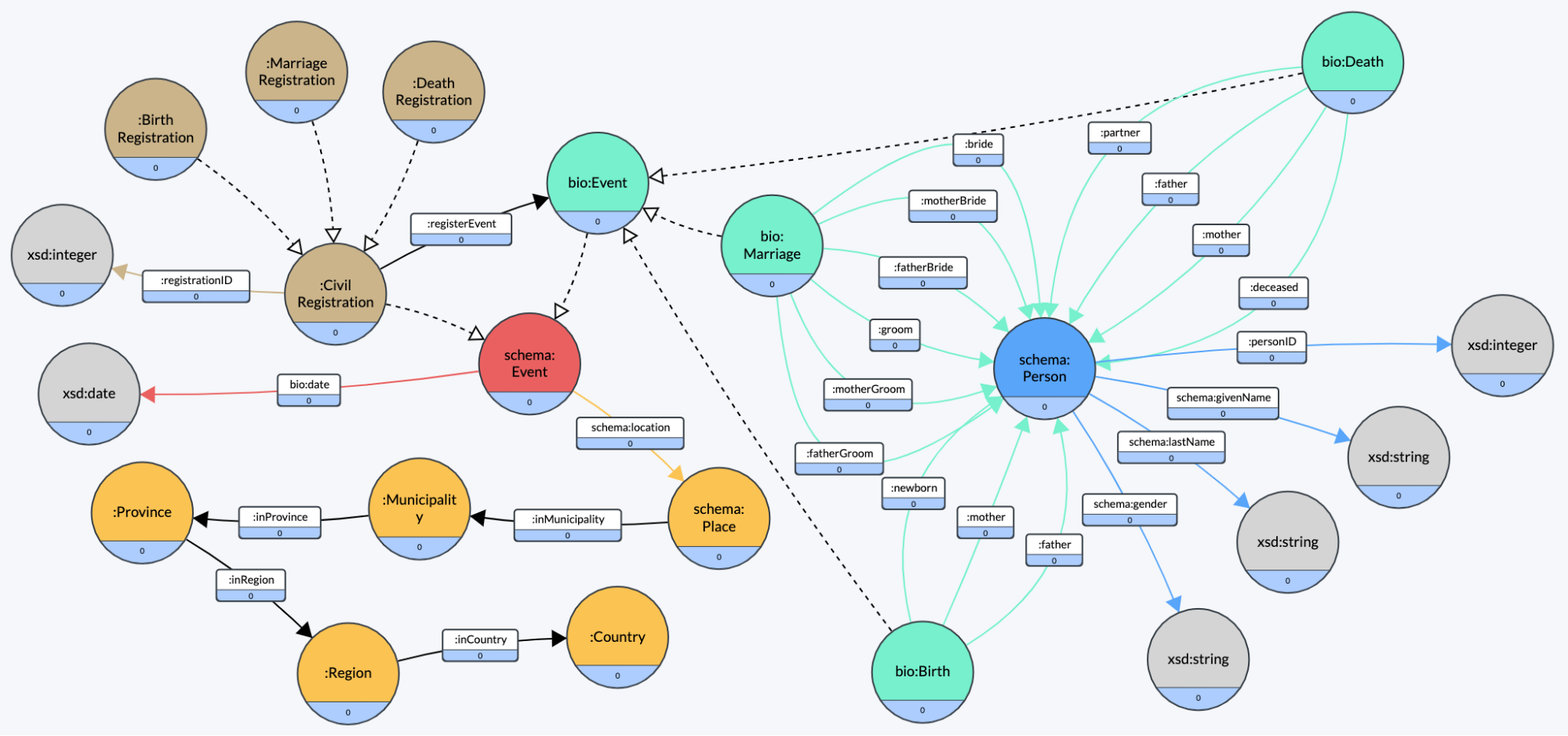}
    \caption{Schema of the LINKS Knowledge Graph, using the \texttt{bio}, \texttt{schema}, \texttt{links} namespaces.}
    \label{fig:links}
\end{figure}

Mentions of historical occupations and religions can happen in datasets such as \textbf{civil registries.} However, an important problem with civil registers deals with their interlinkage: finding the matching birth, marriage, and death certificates of individuals. This is, in fact, a population reconstruction effort with many challenges, such as limited observational data, migration, spelling mistakes, acknowledgement of children, re-use of deceased sibling names, and so forth \cite{mandemakers2017links}. 
Building on the LINKS project\footnote{See \url{https://iisg.amsterdam/en/hsn/projects/links}}
we link the appearance of the same person in 27.5 million birth (1812--1919), marriage (1812--1944) and death (1812--1969) certificates in the Dutch province of Zeeland \cite{raad2020linking}. 
To model the civil registries data, we designed a simple ontology (Figure \ref{fig:links}) that reuses, whenever possible, existing vocabularies such as Schema.org \cite{guha2016schema} and the Bio vocabulary \cite{davis2004bio}. 
The ontology models \emph{civil registrations} (birth, marriage, death), \emph{life events}, and involved \emph{individuals} and \emph{locations}.

A different, more aggregated data source for mentions of historical occupations and religions are \textbf{historical censuses} such as  the Dutch historical censuses (1795-1971) \cite{boonstra2007twee}, a population count including \emph{demography}, \emph{occupations}, and \emph{housing}.
The dataset is available as a collection of 507 Excel spreadsheets, containing 2,288 census tables.\footnote{See \url{http://volkstellingen.nl/nl/index.html}} The CEDAR project \cite{merono2017cedar}, now integrated in CLARIAH, produced a Linked Data version of this dataset, publishing more than 6.8 million statistical observations using the RDF Data Cube vocabulary \cite{world2014rdf}
In addition, we engineered the \texttt{tablinker} vocabulary \cite{meronyo2012}
for the representation of table layout in RDF; this was needed in order to: (a) provide historians with a mechanism for authenticity and interpretability; and (b) enable machines to interpret correctly the roles of table cells and produce a coherent knowledge graph. Other namespaces we use are \texttt{oa} \cite{sanderson2013open} and \texttt{prov} \cite{lebo2013prov}.

\vspace{-.4cm}
\subsubsection{Tools}
\label{sec:ontologies-wp4-tools}


The first step for increasing reuse of our proposed ontologies is to publish them in a \textbf{directory of Humanities ontologies (“Awesome humanities ontologies”\footnote{See \url{https://github.com/CLARIAH/awesome-humanities-ontologies}}).} 
We started this initiative to build a sustainable repository of ontologies for Digital Humanities, utilising the distributed version control system git and the GitHub portal as underlying infrastructure. 
So far, four contributors have committed 23 links to ontologies and vocabularies in 5 categories.

These ontologies can then be used to model any historical structured dataset and convert it into Linked Data with \textbf{COW}.\footnote{See \url{https://pypi.org/project/cow-csvw/}}  
COW is an efficient library and CLI for converting CSV files to RDF, compliant with the CSV on the Web vocabulary \cite{world2016csv}.
CSV files, especially in social economic history, can be arbitrarily large and small and require a mapping to current ontologies and vocabularies to describe the data they contain. 
COW incorporates references to all ontologies surveyed in the previous subsection. 
Users can compose so-called CSVW schema files, where they can specify mappings between the headers and contents of their CSV files, and the classes and properties defined in these ontologies, among other options. The Linked Data that results from this process is therefore highly interoperable by means of reusing these ontologies across various datasets.

\begin{figure}[ht!]
    \centering
    \includegraphics[width=\textwidth]{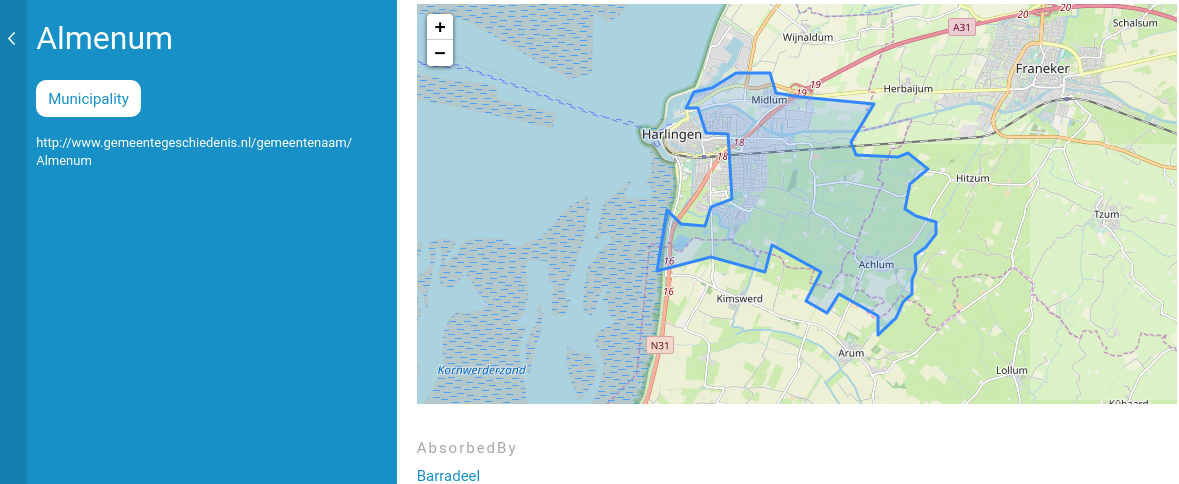}
    \caption{Druid’s browser recognises ontology classes and properties that require specific interaction and rendering components, like maps and images.}
    \label{fig:browser}
\end{figure}

\begin{figure}[ht!]
    \centering
    \subfloat{
    \includegraphics[width=0.5\textwidth]{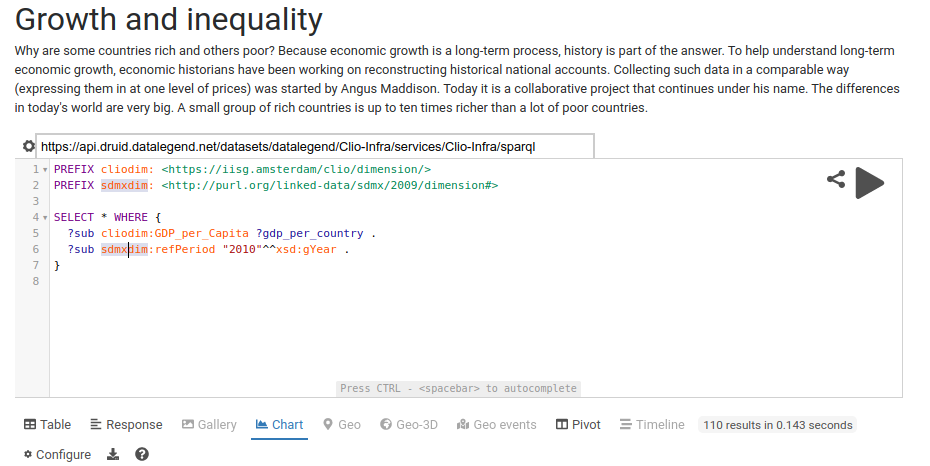}
    }%
    \subfloat{
    \includegraphics[width=0.5\textwidth]{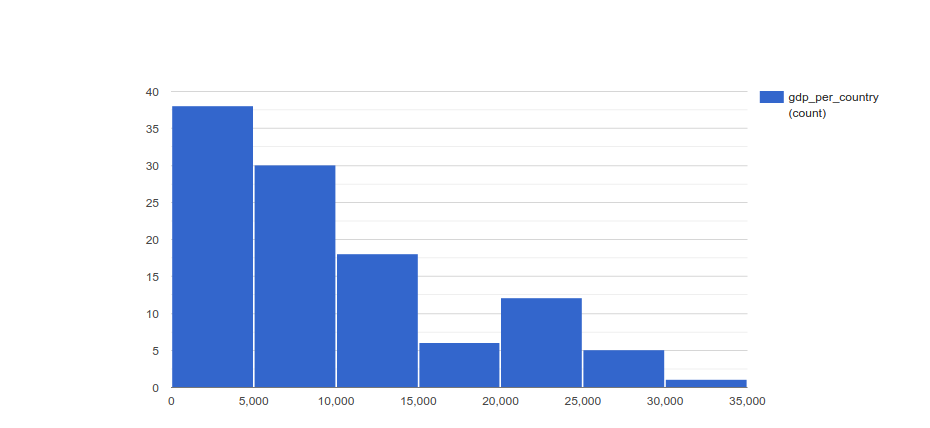}
    }
    \caption{Use of ontologies in queries of a CLARIAH Data story.}
    \label{fig:stories}
\end{figure}

Once created, Linked Data needs to be stored efficiently for publishing and use, and this is what we achieve with \textbf{Druid}.\footnote{See \url{https://druid.datalegend.net/}} Druid is a state-of-the-art, highly scalable triplestore by Triply\footnote{Triply is a spin-off company from CLARIAH started up by former CLARIAH researchers from the Vrije Universiteit Amsterdam; see \url{https://triply.cc/}} based on the HDT (header dictionary triples) technology \cite{fernandez2013binary}. 
The use cases and ontologies of CLARIAH served as an incubator for its design.
In Druid, the ontologies of the previous section play a primary role in diverse Linked Data browsing and interaction tasks. For example, Druid recognises certain classes, properties and values such as polygons and pictures (e.g. the \texttt{og:hasGeometry} and \texttt{og:AsWKT} properties), and adequately renders them in the browser (see Figure \ref{fig:browser}).
Similarly, but operating outside of the triplestore, \textbf{Data stories\footnote{See \url{https://stories.datalegend.net/}}}
aims to provide users with a means to narrate stories based on their Linked Data, combining 
text directly written by users with visualisations resulting from SPARQL queries (see Figure \ref{fig:stories}).
However, rather than creating visualizations and stories, sometimes users are more interested in giving access directly to data through APIs; this is the main purpose of \textbf{grlc}~\cite{merono2016grlc}.\footnote{See \url{http://grlc.io/} and \url{https://github.com/CLARIAH/grlc}} grlc leverages the social context of shared SPARQL queries, such as the endpoint they are directed to, their parameters, and their human-readable descriptions; to automatically generate Linked Data APIs. These APIs can be easily reused by humans and machines to access a large variety of Linked Data sources and Knowledge Graphs. 

\subsection{Linguistics \& Language Technology}
\label{sec:ontologies-wp3}

Many sources relevant to humanities scholars are in textual form \cite{jockers2015text}. The linguistics pillar in CLARIAH  focused on developing tools for linguistic analysis. The different subprojects 
can be divided into two categories: technology for linguistics, and computational linguistics for other humanities disciplines. The latter category includes for example tools to semantically analyse historical texts. 


\subsubsection{Ontologies}
\label{sec:ontologies-wp3-onto}

Within linguistics research, many standards exist to express enrichments of texts. Even for a fairly standardised task such as part-of-speech tagging,\footnote{A task concerned with assigning types such as noun or verb to words} dozens of tagsets exist \cite{petrov2011universal}.  Some of these tagsets only discern the base categories of words, others also include other grammatical information such as the case, number or tense. For other enrichment layers, such as named entities and semantic roles, also various standards exist. It is not the purpose of CLARIAH to reconcile all these standards, but to provide a shared means to work across 
them. The connection to core semantic web ontologies is less strongly developed than in the social and economic history structured data pillar, for two reasons: 1) the scope of the linguistics pillar was broader, with more partners; and 2) more ‘legacy’ tools and resources that needed to be adapted.

Within CLARIAH,
we have started modelling diachronic lexicons as Linked Data to further their interoperability and use in different projects. Diachronic lexicons are dictionaries that describe how the meaning of certain terms has changed over time. Various CLARIAH partners, such as Meertens Institute\footnote{\url{https://www.meertens.knaw.nl/cms/nl/}} and the Dutch Language Institute,\footnote{\url{https://ivdnt.org/the-dutch-language-institute}} have a long history of creating and maintaining such dictionaries, but these did not always find users outside linguistics. By adopting a Linked Data approach, terminological knowledge can be connected more easily to non-terminological resources, adding context from a linguistic perspective. 

\begin{figure}[ht!]
    \centering
    \includegraphics[width=\textwidth]{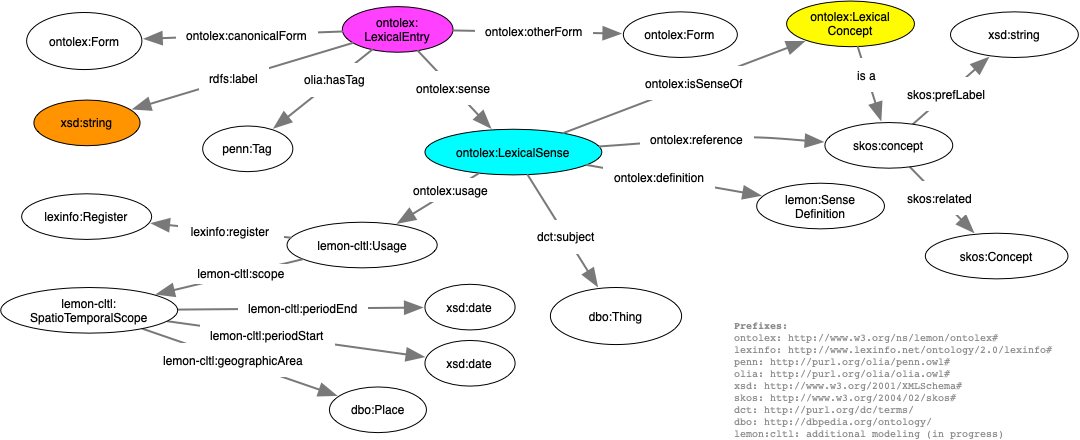}
    \caption{The CLARIAH diachronic lexicon model}
    \label{fig:lexical}
\end{figure}

As a starting point for modelling diachronic linked data, six lexicons were converted \cite{maks2016integrating}. These lexicons cover various topics such as plant names (Pland\footnote{\url{https://www.meertens.knaw.nl/pland/}}) and embodied emotions,\footnote{\url{https://emotionsandsenses.wordpress.com/category/embodied-emotions-2/}} but through evolution in meaning of concepts described in these lexicons shared contexts can be identified. Furthermore, the aim of this exercise was also to show that these different lexicons could be described using a shared data model. 
The model uses \textbf{Lemon - Lexicon Model for Ontologies} \cite{mccrae2017ontolex} where possible and extended with concepts necessary to describe the temporal scope of a word sense. Figure \ref{fig:lexical} represents the CLARIAH diachronic lexicon model. 

\cite{depuydt2018diachronic} further developed diachronic LOD lexicons by the conversion of \textbf{DiaMaNT}, a digital historical language infrastructure. Where in \cite{maks2016integrating} lexical senses were assigned to time periods, this work aggregates senses through attestations, or observed use of the language. This provides an additional etymological provenance trail for the linguistic context of terms. 

\subsubsection{Tools}
\label{sec:ontologies-wp3-tools}

Text enrichments are typically encoded through some form of XML annotation \cite{wilcock2009introduction}. Many different formats are used among different research initiatives in computational linguistics, and some efforts have focused on aligning or mapping between different formats (cf. \cite{hellmann2013integrating}). 

\begin{figure}[ht!]
    \centering
    \includegraphics[width=0.7\textwidth]{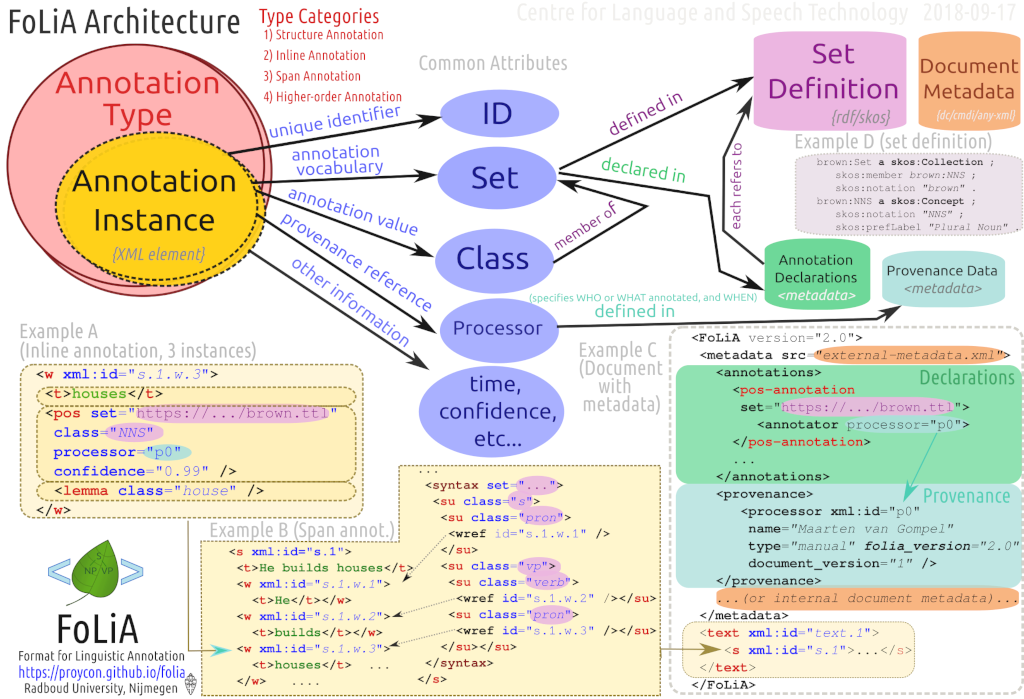}
    \caption{FoLiA architecture (source: \url{https://proycon.github.io/folia/}) }
    \label{fig:annotations}
\end{figure}

Within the CLARIAH project, two linguistic annotation formats are used. These formats were developed in the context of prior research projects and were adopted into the CLARIAH infrastructure because they are the predominant formats in the text analysis tools adopted and further developed in CLARIAH. The main reason to discuss linguistic standards here, is that they provide the bridge between the raw data (text) and semantic layers that connect these data to other (LOD) resources. 
The \textbf{NAF format\footnote{\url{http://wordpress.let.vupr.nl/naf/}}} is a stand-off XML annotation format that combines strengths of the Linguistic Annotation Framework (LAF) and the NLP Interchange Format \cite{fokkens2014naf}. It was developed at the Vrije Universiteit Amsterdam in the BiographyNet and the NewsReader projects (in collaboration with Fondazione Bruno Kessler in Italy and the University of the Basque Country in Spain). It was developed from a need to connect semantic annotation layers to linguistic annotation layers, and has a serialisation to RDF for ingestion into a knowledge graph. 
The \textbf{FoLiA format\footnote{\url{https://proycon.github.io/folia/} Last visited: 28 July 2020.}} is a mixture of an inline and stand-off XML annotation format. It was first and foremost developed with the focus of storing and exchanging textual resources (including large corpora) in mind \cite{gompel2017folia}. Its development started in the context of CLARIN-NL,\footnote{\url{https://www.clarin.nl/}} DutchSemCor,\footnote{\url{http://wordpress.let.vupr.nl/dutchsemcor/}} and OpenSoNaR\footnote{\url{https://portal.clarin.nl/node/4195}} at Tilburg University and was continued at Radboud University. Figure \ref{fig:annotations} provides an overview of the FoLiA architecture. 
Within the CLARIAH project, a \textbf{conversion tool\footnote{\url{https://github.com/cltl/NAFFoLiAPy} Last visited: 28 July 2020}} between NAF and FoLiA was created. However, as new features are still added to each of these formats, the formats are sometimes out of sync in what they can encode, and subsequently the tool cannot always convert all layers of one format into the other. 

It should be noted that NAF and FoLiA are less focused on describing the form of a text, as TEI \cite{ide1995text}
is, but rather the linguistic and semantic characteristics of the words. However, both NAf and FoLiA encode some text shape characteristics such as paragraph, sentence and word boundaries. Mappings and tools exist to convert a subset of TEI to FoLiA. 

\subsection{Multimedia \& Media Studies}
\label{sec:ontologies-wp5}

Media Scholars typically are interested in heterogeneous multimedia data sets, and 
they want to investigate the various sources together \cite{bron2016media},
including the radio, television, amateur material and contextual datasets that shape society.
Within the CLARIAH Media pillar, the main product is the MediaSuite \cite{martinez2017tools}, which bundles multiple forms of search and browsing 
based on enriched textual and structured metadata over multimedia sources, primarily from the Netherlands Institute for Sound and Vision\footnote{\url{https://www.beeldengeluid.nl/en}} (NISV), the Royal Library's newspaper collection, the DANS oral history collections and several Open Images collections.


\subsubsection{Ontologies}
\label{sec:ontologies-wp5-onto}

We reused a number of existing ontologies to model the media collections.
As media scholars often look at (media) events, we needed a model that is able to express a variety of events. The \textbf{Simple Event Model (SEM)} \cite{van2011design} allows for the representation of events, actors, locations and temporal descriptions \cite{van2011design}, but also for relations between, for example, events and persons to express the role that a person played in an event. 
\textbf{SKOS} \cite{miles2005skos} is used  to represent concepts from structured vocabularies, including the GTAA (Common Thesaurus for AudioVisual Heritage\footnote{\url{https://old.datahub.io/dataset/gemeenschappelijke-thesaurus-audiovisuele-archieven}}), which is a core vocabulary in our tool, as it is one specifically to describe (Dutch) audiovisual material and is shared by a number of AV archives in the Netherlands. 
\textbf{Dublin Core} \cite{weibel1998dublin} is used for the descriptive metadata of the objects. 
We use the \textbf{W3C Web Annotation Model} \cite{sanderson2013open} to model user annotations, both from expert users, and resulting from crowdsourcing initiatives. 

In addition to these ontologies, 
from 2006 to 2018 a catalog management system (CMS) was developed within the NISV that ties in with the main processes of registration, digitisation and access to a large number of audiovisual items as a precurosor of the \textbf{NISV ontology}. 
The system, iMMix, had a data model originally derived from FRBR 
\cite{tillett2005frbr}
adapting its hierarchical form
with work, series, season, program and segment. In mid-2018, the NISV switched to a new catalog management system, called DAAN, which is an adapted version of the VizOne product from VizRT. All metadata hierarchies in Viz One use metadata fields based upon the Dublin Core Metadata Initiative (DCMI). Where iMMix was built roughly around the programs, the DAAN model is built around the physical carriers, called items, that can be part of a program, which can be part of a season, a series. The work level has been dropped. The DAAN model also contains the notion of log track items that describe additional, mostly time-coded metadata such as annotations and transcriptions.
The existing NISV data models are modelled in an RDFS \cite{brickley1999resource} scheme. Another approach could be re-using a widely used ontology and expressing the NISV data in that model. The benefit of the current approach is that the RDFS scheme for the NISV data is as close to the original data as possible, while the option remains open to express the classes in a different ontology by using \texttt{rdfs:subClassOf} and the properties using \texttt{rdfs:subPropertyOf}. The NISV thesaurus is already described using the SKOS ontology and therefore the classes that are tied to the controlled vocabularies of the thesaurus are defined as SKOS concepts using the \texttt{rdfs:subClassOf} relation. Further mappings to other desired target models (including schema.org, ebuCore/pbCore or Europeana Data Model \cite{doerr2010europeana}) are planned for.

\begin{figure}[ht!]
    \centering
    \includegraphics[width=0.7\textwidth]{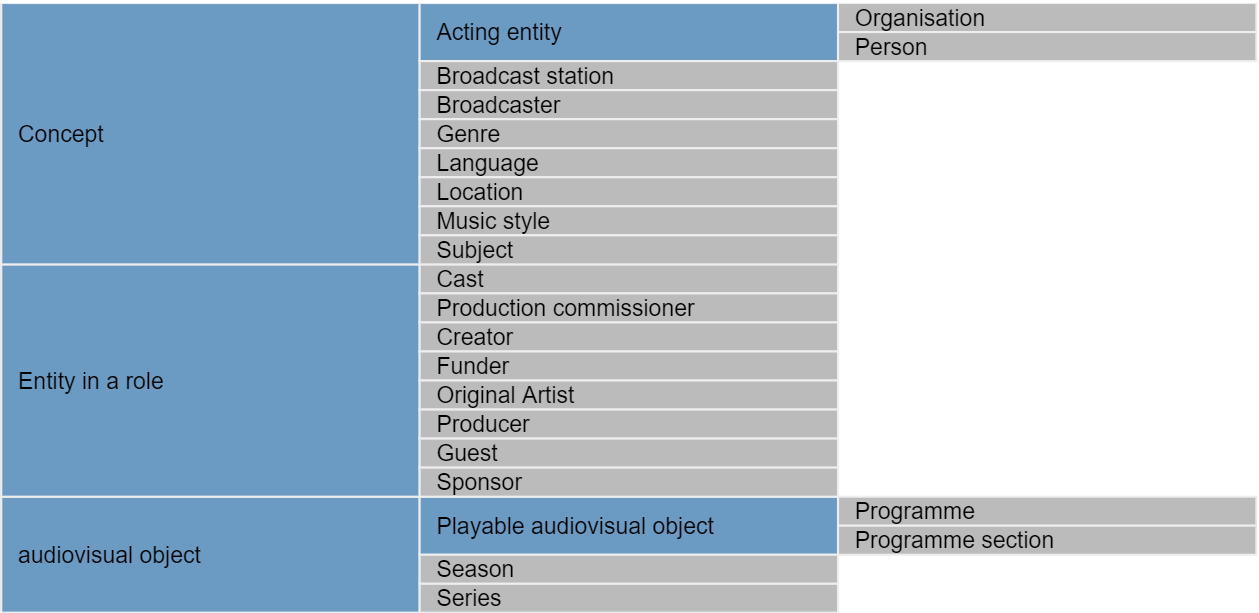}
    \caption{Partition table of the NISV Ontology (image produced with Ontospy)}
    \label{fig:menu}
\end{figure}

\subsubsection{Tools}
\label{sec:ontologies-wp5-tools}

The \textbf{CLARIAH Media Suite} is a research environment and core facility
for CLARIAH multimedia and media studies.
It brings together data and tools in a virtual workspace for researchers interested in (Dutch) media. It facilitates access to key Dutch audio-visual and contextual collections with advanced mixed media search and analysis tools. The Media Suite includes tooling to: (a) inspect data using critical methods and various views on collection metadata; (b) provide faceted search options for distant reading, corpus selection and close reading; and (c) provides an exploratory browser, to facilitate serendipitous discovery. 
This \textbf{Exploratory linked media browser} is based on the DIVE tool, where items from various collections are linked through shared vocabularies and ontologies \cite{de2015dive}. To make an interconnected knowledge graph which can be used for exploratory search, we employ various strategies for enrichment: We establish mappings from collection-specific metadata to generic terms for each of the collections. This ensures that queries on this generic level (such as retrieving a textual description for an item) return relevant results from each of these collections. Using alignment services such as Cultuurlink,\footnote{\url{http://cultuurlink.beeldengeluid.nl/}} we establish correspondences between the persons, places and events found in our enrichments and structured vocabularies \cite{de2017enriching}.

\begin{figure}[ht!]
    \centering
    \includegraphics[width=0.85\textwidth]{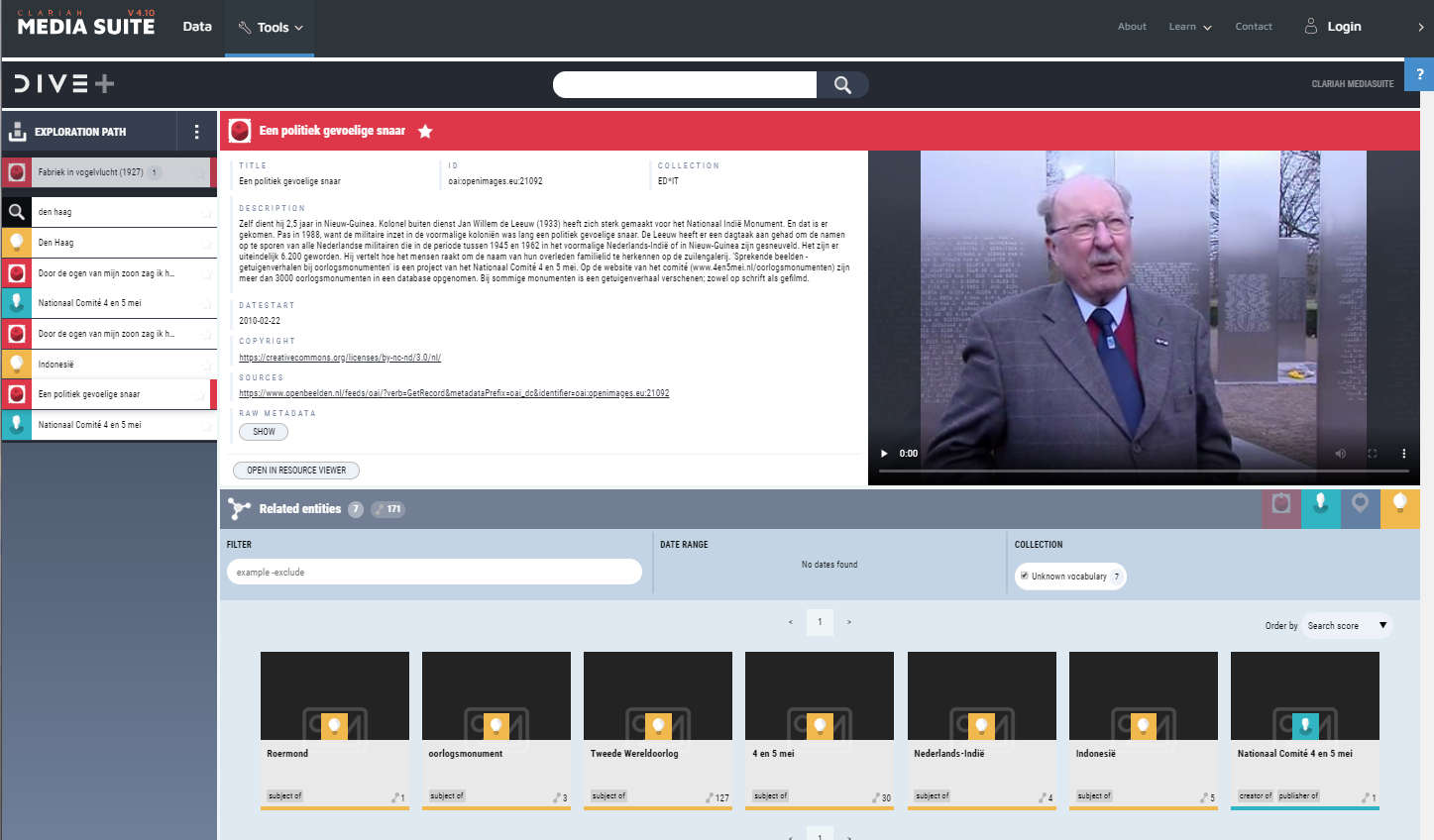}
    \caption{Screenshot of the Exploratory Linked Media Browser as part of the MediaSuite. On the left, a ‘bread crumb’ trail of browsed concepts, persons and media items is shown. Links between the current item (shown in the top) result in suggested items in the bottom.}
    \label{fig:mediasuite}
\end{figure}

\subsection{Inter-pillar Interoperability}
\label{sec:ontologies-inter}

To investigate how the developed ontologies and tools would perform in practice, CLARIAH 
funded 16 research pilot projects of \euro{}60k and a runtime of at most 1 year, and 4 Accelerating Scientific Discovery in the Arts and Humanities (ADAH) projects
in collaboration with the Netherlands eScience center
(\euro{}100k + 1.5 research engineers).
In addition, the Amsterdam Time Machine\footnote{\url{https://amsterdamtimemachine.nl/hisgis-clariah/}} project was funded as an accelerator project that spanned all pillars in connection to the European Time Machine\footnote{\url{https://www.timemachine.eu/}} project. 
These first adopters yielded new ontologies and tools, as well as insights into strengths and limitations.


\subsubsection{Ontologies}
\label{sec:ontologies-inter-onto}

In the \textbf{CLARIAH Amsterdam Time Machine}, three use cases focusing on Amsterdam from each CLARIAH pillar were connected through a historical geographical infrastructure to answer questions of inequality and cultural customs in specific places and times.
The difficulty in this and any other historical urban project, is the fact that addresses cannot be used as unique identifiers as they change over time.
There are various ways to model such changes \cite{beek2018nlgis}, and we chose to use the space-time-prism model \cite{kessler2015querying}, because their model exemplified via CO2 measurements perfectly fitted our use case. Just as with the CO2 measurements, where there is a certain ‘ping’ at which time, location and measurement value are observed, the HISGIS project specified ‘pings’ via a fixed set of location points (that can be expanded on by others) 
\cite{raat2019hisgis}. The location points have a geographical representation, a time indication (year), and the address as value. Hence, sources mentioning this value can be directly linked to the location point, allowing for multiple datasets to be combined. For example, in the linguistics use case, Amsterdam dialects were investigated, for which the diachronic lexicon model Lemon (Section \ref{sec:ontologies-wp3}) was used \cite{maks2016integrating}. Through the space-time-prism model, we were able to study to what extent there was overlap in the social and economic position of individuals (modeled via HISCO and HISCAM) and the social economic prestige of Amsterdam dialects \cite{nederlandse2020}.

The basis for the CLARIAH diachronic lexicon model \cite{depuydt2018diachronic} was also reused in the \textbf{Diamonds in Borneo} project \cite{hofmeester2019diamonds}. In this project, the causes and consequences of the circulation of people, commodities and ideas in a globalising world are investigated through the automatic analysis of large historical newspaper corpus from the Dutch National Library. 

\subsubsection{Tools}
\label{sec:ontologies-inter-tools}

Various research pilots and ADAH projects revolve around text, for which the tools from the linguistics pillar were put to the test. In the \textbf{EviDENce}\footnote{\url{https://www.esciencecenter.nl/projects/evidence/}} project, historians wanted to analyse oral history records and ego documents (e.g. letters, diaries) to trace how the concept of violence changed over time. The project made use of various tools, such as the \textbf{CLARIAH Robust Semantic Parsing pipeline},\footnote{\url{https://github.com/CLARIAH/wp3-semantic-parsing-Dutch} Last visited: 28 July 2020} which for example outputs NAF text annotations and SEM events, in combination with prior processed data from Nederlab formatted in FoLiA. The project revealed that  the fine-grained level of linguistic analysis was too detailed for the research questions \cite{hogervorst2018event}.

\section{Discussion, Guidelines and Conclusion} 
\label{sec:conclusion}

A large degree of standardisation and, concomitantly, interoperability without loss of information is possible with ontologies and vocabularies; but at the cost of relying on a large number of different ontologies. A large number of ontologies can be both a strength and a weakness, as more ontologies can represent a multitude of domains in detail, while more ontologies also imply larger integration costs. In total, we have used 26 different ontologies, taxonomies, classification systems and lexicons. Of these, 6 were engineered from scratch to model new humanities domains, while 20 were reused or converted. Many such reused ontologies and vocabularies, like RDFS, OWL, PROV, SKOS and Schema.org, apply not just to the humanities but to broader contexts; while others, such as Bio, IPUMS, HISCO and HL7 are more 
specific to humanities domains.
But, despite this reuse, our ecosystem is still highly fragmented, which signals the difficult challenge of collaborative ontology engineering even in highly controlled environments. 

To address the issue of using generic ontologies where possible, and domain ontologies where needed, we combined bottom-up and top-down approaches where the CLARIAH community was central. 
First, the showcase of technical demonstrations in workshops and community days pushed the community's enthusiasm into the possibilities of linking data across different sources. But linking more data soon derived into various models that partly overlapped and needed to be aligned. Second, and to address this, we found Schema.org to be a common denominator in various projects,
and hence we encouraged its reuse as a powerful tool for ontological cooperation. This was due to a number of reasons: 
(a) 
a general schema
a higher chance of being picked and used by large search indexes, enabling their findability and reuse
in 
tools like e.g. Google Dataset Search \cite{brickley2019google}; (b) one central vocabulary that covers a large portion 
of a community’s modelling needs feels more convenient to users; (c) even if that coverage is not perfect, such vocabularies might add the required little semantics to go a long way \cite{hendler2013little}; and (d) these vocabularies can be extended 
to reach a satisfactory specialisation level.
Therefore, prioritising vocabularies seems to be the best solution to the necessity of the different ontologies. 
Lastly, we also verified that the need for domain specific ontologies remained ever-growing: this led to using the ``Awesome humanities ontologies'' and Linked Open Vocabularies \cite{vandenbussche2017linked} lists, again community-based efforts. 




In this chapter, we summarised the efforts within the CLARIAH project to achieve interoperability and enable broadly quantitative research in linguistics, social and economic history, media studies, and the humanities as a whole. To this end Semantic Web based ontologies, tools, cross-dataset queries, and community initiatives (e.g. “awesome humanities ontologies”) for the humanities were developed. 
By developing
these
decentralised, yet controlled Knowledge Graph development practices we have contributed to increasing interoperability in the humanities
and 
enabling new research opportunities to a wide range of scholars. However, we observe that users without Semantic Web knowledge find these technologies hard to use, and place high value in end-user tools
that enable engagement. Therefore, for the future we emphasise the importance of tools to specifically target the goals of concrete communities --in our case, the analytical and quantitative answering of humanities research questions for humanities scholars. In this sense, usability is not just important in a tool context; in our view, we need to empower users in deciding under what models these tools operate.

\bibliographystyle{splncs04}
\bibliography{bibliography.bib}

\begin{thebibliography}{10}
\providecommand{\url}[1]{\texttt{#1}}
\providecommand{\urlprefix}{URL }
\providecommand{\doi}[1]{https://doi.org/#1}

\bibitem{adamou2016whise}
Adamou, A., Daga, E., Isaksen, L.: {WHiSe 2016, Humanities in the Semantic Web.
  Proceedings of the 1st Workshop on Humanities in the Semantic Web, co-located
  with 13th ESWC Conference 2016 (ESWC 2016)}. CEUR-WS.org (2016)

\bibitem{adamou2017whise}
Adamou, A., Daga, E., Isaksen, L.: {WHiSe 2017, Workshop on Humanities in the
  Semantic Web. Proceedings of the Second Workshop on Humanities in the
  Semantic Web (WHiSe II), co-located with 16th International Semantic Web
  Conference (ISWC 2017)}. CEUR-WS.org (2017)

\bibitem{ashkpour2019theory}
Ashkpour, A.: Theory and Practice of Historical Census Data Harmonization: The
  Dutch historical census use case: a flexible, structured and accountable
  approach using Linked Data technology. Ph.D. thesis, Erasmus University
  Rotterdam (2019)

\bibitem{beek2018nlgis}
Beek, W., Zijdeman, R.: {nlGis: A use case in linked historic geodata}. In:
  European Semantic Web Conference. pp. 437--447. Springer (2018)

\bibitem{boonstra2007twee}
Boonstra, O., Doorn, P., van Horik, M., van Maarseveen, J., Oudhof, J.: {Twee
  eeuwen Nederland geteld. Onderzoek met de digitale Volks-, Beroeps-en
  Woningtellingen 1795-2001}. In: DANS Symposium publicaties; 2. Den Haag: DANS
  (2007)

\bibitem{brickley2019google}
Brickley, D., Burgess, M., Noy, N.: Google dataset search: Building a search
  engine for datasets in an open web ecosystem. In: The World Wide Web
  Conference. pp. 1365--1375 (2019)

\bibitem{brickley1999resource}
Brickley, D., Guha, R.V., Layman, A.: {Resource description framework (RDF)
  schema specification}. Tech. rep., W3C (1999)

\bibitem{bron2016media}
Bron, M., Van~Gorp, J., De~Rijke, M.: {Media studies research in the
  data-driven age: How research questions evolve}. Journal of the Association
  for Information Science and Technology  \textbf{67}(7),  1535--1554 (2016)

\bibitem{collins2003human}
Collins, F.S., Morgan, M., Patrinos, A.: The human genome project: lessons from
  large-scale biology. Science  \textbf{300}(5617),  286--290 (2003)

\bibitem{world2014rdf}
Cyganiak, R., Reynolds, D.: {The RDF data cube vocabulary}. Tech. rep., World
  Wide Web Consortium (2014)

\bibitem{davis2004bio}
Davis, I., Galbraith, D.: {BIO: A vocabulary for biographical information}.
  Vocab.org  (2004)

\bibitem{de2017enriching}
De~Boer, V., Melgar, L., Inel, O., Ortiz, C.M., Aroyo, L., Oomen, J.: Enriching
  media collections for event-based exploration. In: Research Conference on
  Metadata and Semantics Research. pp. 189--201. Springer (2017)

\bibitem{de2015dive}
De~Boer, V., Oomen, J., Inel, O., Aroyo, L., Van~Staveren, E., Helmich, W.,
  De~Beurs, D.: {DIVE into the event-based browsing of linked historical
  media}. Journal of Web Semantics  \textbf{35},  152--158 (2015)

\bibitem{depuydt2018diachronic}
Depuydt, K., De~Does, J.: {The diachronic semantic lexicon of dutch as linked
  open data}. In: Proceedings of the Eleventh International Conference on
  Language Resources and Evaluation (LREC 2018). European Language Resources
  Association (ELRA), Paris, France (2018)

\bibitem{doerr2003cidoc}
Doerr, M.: {The CIDOC conceptual reference module: an ontological approach to
  semantic interoperability of metadata}. AI magazine  \textbf{24}(3),  75--75
  (2003)

\bibitem{doerr2010europeana}
Doerr, M., Gradmann, S., Hennicke, S., Isaac, A., Meghini, C., Van~de Sompel,
  H.: The europeana data model (edm). In: World Library and Information
  Congress: 76th IFLA general conference and assembly. vol.~10, p.~15 (2010)

\bibitem{elias1997occupational}
Elias, P.: {Occupational classification (ISCO-88): Concepts, methods,
  reliability, validity and cross-national comparability}  (1997)

\bibitem{fernandez2013binary}
Fern{\'a}ndez, J.D., Mart{\'\i}nez-Prieto, M.A., Guti{\'e}rrez, C., Polleres,
  A., Arias, M.: {Binary RDF representation for publication and exchange
  (HDT)}. Journal of Web Semantics  \textbf{19},  22--41 (2013)

\bibitem{fokkens2014naf}
Fokkens, A., Soroa, A., Beloki, Z., Ockeloen, N., Rigau, G., Van~Hage, W.R.,
  Vossen, P.: {NAF and GAF: Linking linguistic annotations}. In: Proceedings
  10th Joint ISO-ACL SIGSEM Workshop on Interoperable Semantic Annotation. pp.
  9--16 (2014)

\bibitem{gangemi2009ontology}
Gangemi, A., Presutti, V.: Ontology design patterns. In: Handbook on
  ontologies, pp. 221--243. Springer (2009)

\bibitem{garijo2014common}
Garijo, D., Alper, P., Belhajjame, K., Corcho, O., Gil, Y., Goble, C.: Common
  motifs in scientific workflows: An empirical analysis. Future Generation
  Computer Systems  \textbf{36},  338--351 (2014)

\bibitem{gompel2017folia}
Gompel, M.v., Sloot, K., Reynaert, M., van~den Bosch, A.: {FoLiA in Practice.
  The Infrastructure of a Linguistic Annotation Format}. CLARIN-NL in the Low
  Countries  (2017)

\bibitem{guha2016schema}
Guha, R.V., Brickley, D., Macbeth, S.: {Schema.org: Evolution of Structured
  data on the Web}. Communications of the ACM  \textbf{59}(2),  44--51 (2016)

\bibitem{haigh2014we}
Haigh, T.: We have never been digital. Communications of the ACM
  \textbf{57}(9),  24--28 (2014)

\bibitem{heath2011linked}
Heath, T., Bizer, C.: Linked data: Evolving the web into a global data space.
  Synthesis lectures on the semantic web: theory and technology  \textbf{1}(1),
   1--136 (2011)

\bibitem{hellmann2013integrating}
Hellmann, S., Lehmann, J., Auer, S., Br{\"u}mmer, M.: {Integrating NLP using
  linked data}. In: International semantic web conference. pp. 98--113.
  Springer (2013)

\bibitem{hendler2013little}
Hendler, J.: A little semantics goes a long way. URL: http://www. cs. rpi.
  edu/\~{} hendler/LittleSemanticsWeb. html [November 24, 2014]  (2013)

\bibitem{hoekstra2018datalegend}
Hoekstra, R., Mero{\~n}o-Pe{\~n}uela, A., Rijpma, A., Zijdeman, R., Ashkpour,
  A., Dentler, K., Zandhuis, I., Rietveld, L.: {The dataLegend ecosystem for
  historical statistics}. Journal of Web Semantics  \textbf{50},  49--61 (2018)

\bibitem{hofmeester2019diamonds}
Hofmeester, K., Ashkpour, A., Depuydt, K., de~Does, J.: {Diamonds in Borneo:
  Commodities as Concepts in Context}. In: Proceedings of the 3rd International
  Conference on Digital Access to Textual Cultural Heritage. pp. 45--50 (2019)

\bibitem{hogervorst2018event}
Hogervorst, S., Brugman, H., Buitinck, L., van Erp, M., Klijn, E., Kouw, W.,
  de~Vos, M., Willemsen, J.: {The Event-Detection GAP: Manual vs. automatic
  event detection in historical research}. In: DHBenelux (2018)

\bibitem{ide1995text}
Ide, N., V{\'e}ronis, J.: Text encoding initiative: Background and contexts,
  vol.~29. Springer Science \& Business Media (1995)

\bibitem{nederlandse2020}
Instituut, M.: {Nederlandse Dialectenbank}.
  \url{https://www.meertens.knaw.nl/ndb/} (2020), date of last Access: April 3,
  2020

\bibitem{jockers2015text}
Jockers, M.L., Underwood, T.: {Text-mining the humanities}. A new companion to
  digital humanities pp. 291--306 (2015)

\bibitem{kessler2015querying}
Ke{\ss}ler, C., Farmer, C.J.: Querying and integrating spatial--temporal
  information on the web of data via time geography. Journal of Web Semantics
  \textbf{35},  25--34 (2015)

\bibitem{lambert2013construction}
Lambert, P.S., Zijdeman, R.L., Van~Leeuwen, M.H., Maas, I., Prandy, K.: {The
  construction of HISCAM: A stratification scale based on social interactions
  for historical comparative research}. Historical Methods: A Journal of
  Quantitative and Interdisciplinary History  \textbf{46}(2),  77--89 (2013)

\bibitem{lebo2013prov}
Lebo, T., Sahoo, S., McGuinness, D., Belhajjame, K., Cheney, J., Corsar, D.,
  Garijo, D., Soiland-Reyes, S., Zednik, S., Zhao, J.: {PROV-O: The PROV
  Ontology}. W3C recommendation  \textbf{30} (2013)

\bibitem{leeuwen2002hisco}
Leeuwen, M.v., Maas, I., Miles, A.: {HISCO: Historical international standard
  classification of occupations}. Leuven: Leuven University Press (2002)

\bibitem{maks2016integrating}
Maks, I., van Erp, M., Vossen, P., Hoekstra, R., van~der Sijs, N.: {Integrating
  diachronous conceptual lexicons through linked open data} (2016)

\bibitem{mandemakers2017links}
Mandemakers, K., Laan, F.: {LINKS Dataset Genes Germs and Resources}. WieWasWie
  Zeeland. Civil Certificates  (2017)

\bibitem{martinez2017tools}
Martinez-Ortiz, C., Ordelman, R., Koolen, M., Noordegraaf, J., Melgar, L.,
  Aroyo, L., Blom, J., van Gorp, J., Baaren, E., Beelen, K., et~al.: From tools
  to ``recipes”: Building a media suite within the dutch digital humanities
  infrastructure clariah. In: DH Benelux (2017)

\bibitem{mccrae2017ontolex}
McCrae, J.P., Bosque-Gil, J., Gracia, J., Buitelaar, P., Cimiano, P.: {The
  Ontolex-Lemon model: development and applications}. In: Proceedings of eLex
  2017 conference. pp. 19--21 (2017)

\bibitem{merono2017cedar}
Mero{\~n}o-Pe{\~n}uela, A., Ashkpour, A., Gu{\'e}ret, C., Schlobach, S.:
  {CEDAR: the Dutch historical censuses as linked open data}. Semantic Web
  \textbf{8}(2),  297--310 (2017)

\bibitem{meronyo2012}
Mero{\~n}o-Pe{\~n}uela, A., Ashkpour, A., Rietveld, L., Hoekstra, R.,
  Schlobach, S.: {Linked Humanities Data: The Next Frontier? A Case-study in
  Historical Census Data}. In: {Proceedings of the 2nd International Workshop
  on Linked Science (LISC2012). International Semantic Web Conference (ISWC)}.
  vol.~951. CEUR Workshop Proceedings (2012), \url{http://ceur-ws.org/Vol-951/}

\bibitem{merono2015semantic}
Mero{\~n}o-Pe{\~n}uela, A., Ashkpour, A., Van~Erp, M., Mandemakers, K., Breure,
  L., Scharnhorst, A., Schlobach, S., Van~Harmelen, F.: Semantic technologies
  for historical research: A survey. Semantic Web  \textbf{6}(6),  539--564
  (2015)

\bibitem{merono2016grlc}
Mero{\~n}o-Pe{\~n}uela, A., Hoekstra, R.: {grlc Makes GitHub Taste Like Linked
  Data APIs}. In: European Semantic Web Conference. pp. 342--353. Springer
  (2016)

\bibitem{miles2005skos}
Miles, A., Matthews, B., Wilson, M., Brickley, D.: {SKOS core: Simple Knowledge
  Organisation for the Web}. In: International Conference on Dublin Core and
  Metadata Applications. pp. 3--10 (2005)

\bibitem{noy2009bioportal}
Noy, N.F., Shah, N.H., Whetzel, P.L., Dai, B., Dorf, M., Griffith, N., Jonquet,
  C., Rubin, D.L., Storey, M.A., Chute, C.G., et~al.: Bioportal: ontologies and
  integrated data resources at the click of a mouse. Nucleic acids research
  \textbf{37}(suppl\_2),  W170--W173 (2009)

\bibitem{petrov2011universal}
Petrov, S., Das, D., McDonald, R.: {A universal part-of-speech tagset}. arXiv
  preprint arXiv:1104.2086  (2011)

\bibitem{raad2020linking}
Raad, J., Mourits, R., Rijpma, A., Schalk, R., Zijdeman, R., Mandemakers, K.,
  Merono-Penuela, A.: {Linking Dutch Civil Certificates}. In: WHiSe@ ESWC
  (2020)

\bibitem{raat2019hisgis}
Raat, M., Zijdeman, R.: {HISGIS Amsterdam Location Points, IISH Data
  Collection, V3}. \url{https://hdl.handle.net/10622/FHJJYK} (2019)

\bibitem{renckens2016digital}
Renckens, E.: Digital humanities verfrissen onze blik op bestaande data. E-Data
  \& Research  \textbf{10} (2016)

\bibitem{ruggles2019history}
Ruggles, S., Magnuson, D.L.: {The History of Quantification in History: The JIH
  as a Case Study}. Journal of Interdisciplinary History  \textbf{50}(3),
  363--381 (2019)

\bibitem{sanderson2013open}
Sanderson, R., Ciccarese, P., Van~de Sompel, H., Bradshaw, S., Brickley, D.,
  a~Castro, L.J.G., Clark, T., Cole, T., Desenne, P., Gerber, A., et~al.: Open
  annotation data model. W3C community draft  (2013)

\bibitem{world2016csv}
Tennison, J.: {CSV on the Web: A Primer}. Tech. rep., World Wide Web Consortium
  (2016)

\bibitem{tillett2005frbr}
Tillett, B.: {What is FRBR? A conceptual model for the bibliographic universe}.
  The Australian Library Journal  \textbf{54}(1),  24--30 (2005)

\bibitem{van2011design}
Van~Hage, W.R., Malais{\'e}, V., Segers, R., Hollink, L., Schreiber, G.:
  {Design and use of the Simple Event Model (SEM)}. Journal of Web Semantics
  \textbf{9}(2),  128--136 (2011)

\bibitem{vandenbussche2017linked}
Vandenbussche, P.Y., Atemezing, G.A., Poveda-Villal{\'o}n, M., Vatant, B.:
  {Linked Open Vocabularies (LOV): a gateway to reusable semantic vocabularies
  on the Web}. Semantic Web  \textbf{8}(3),  437--452 (2017)

\bibitem{weibel1998dublin}
Weibel, S., Kunze, J., Lagoze, C., Wolf, M.: Dublin core metadata for resource
  discovery. Internet Engineering Task Force RFC  \textbf{2413}(222), ~132
  (1998)

\bibitem{wilcock2009introduction}
Wilcock, G.: {Introduction to linguistic annotation and text analytics}.
  Synthesis Lectures on Human Language Technologies  \textbf{2}(1),  1--159
  (2009)

\bibitem{wilkinson2016fair}
Wilkinson, M.D., Dumontier, M., Aalbersberg, I.J., Appleton, G., Axton, M.,
  Baak, A., Blomberg, N., Boiten, J.W., da~Silva~Santos, L.B., Bourne, P.E.,
  et~al.: The fair guiding principles for scientific data management and
  stewardship. Scientific data  \textbf{3}(1), ~1--9 (2016)

\end{thebibliography}

\end{document}